\documentclass{article}

\usepackage{times}
\usepackage{graphicx} 
\usepackage{subfigure}

\usepackage{natbib}

\usepackage{amsmath,amssymb}
\usepackage{algorithm}
\usepackage{algorithmic}
\usepackage{tikz}

\usepackage{hyperref}

\usepackage[accepted]{icml2015}

\icmltitlerunning{Deep Clustered Convolutional Kernels}

\begin{document} 

\twocolumn[
\icmltitle{Deep Clustered Convolutional Kernels}


\icmlauthor{Minyoung Kim}{minyoung.kim@us.panasonic.com}
\icmlauthor{Luca Rigazio}{luca.rigazio@us.panasonic.com}
\icmladdress{Panasonic Silicon Valley Laboratory \\
             10900 North Tantau Avenue, Cupertino, California 95014}

\icmlkeywords{Deep Neural Networks, Convolutional Neural Networks, DNN, CNN, Clustering, SGD}

\vskip 0.3in
]

\begin{abstract}
Deep neural networks have recently achieved state of the art performance thanks to new training algorithms for rapid parameter estimation and new regularization methods to reduce overfitting. However, in practice the network architecture has to be manually set by domain experts, generally by a costly trial and error procedure, which often accounts for a large portion of the final system performance. 
We view this as a limitation and propose a novel training algorithm that automatically optimizes network architecture, by progressively increasing model complexity and then eliminating model redundancy by selectively removing parameters at training time. 
For convolutional neural networks, our method relies on iterative split/merge clustering of convolutional kernels interleaved by stochastic gradient descent. 
We present a training algorithm and experimental results on three different vision tasks, showing improved performance compared to similarly sized hand-crafted architectures.
\end{abstract} 

\section{Introduction}
\label{intro}
Recently, deep neural networks (DNNs) have led to significant improvement in several machine learning domains, from speech recognition~\cite{dahl2012context} to computer vision~\cite{krizhevsky2012imagenet,taigman2013deepface} and machine translation~\cite{NIPS2014_5346}. DNNs have reached state of the art performance thanks to their theoretically proven modeling and generalization capabilities~\cite{hornik1989multilayer, hornik1991approximation, kuurkova1992kolmogorov}, and practically driven by improvements in training algorithms for rapid parameter estimation~\cite{martens2010deep, sutskever2013importance}, novel regularization methods to reduce overfitting~\cite{JMLR:v15:srivastava14a} as well as ever increasing data-sets~\cite{imagenet_cvpr09} and powerful new computing platforms~\cite{CUDNN}.
However, before parameter estimation (so called training) can begin the DNN's structure (also called model architecture) is usually manually defined by domain experts \cite{NIN}, and can often account for a substantial portion of the final system performance \cite{GoogLeNet}. We view this step as a bottleneck in the current deep-learning pipeline, one that relies on a trial and error human expert in the loop approach which is, to say the least, rather alchemic in nature. We want to address this basic scalability issue of the deep learning development pipeline with training methods that automatically search for DNN architectures while jointly estimating model parameters.

While structural optimization is a notoriously difficult combinatorial task, successful strategies were adopted in the past for (shallow) models that motivated our approach.
For instance, for Hidden Markov Models with Gaussian mixture kernels, split/merge algorithms were used to independently vary model complexity for each HMM state, resulting in improved accuracy for large vocabulary speech recognition~\cite{sankar1998experiments}.
Information theoretic methods, such as the minimum description length criterion, were also applied to the problem of structural optimization \cite{barron1998minimum}, resulting in improved performance in speech recognition~\cite{shinoda2000mdl} and as well as training algorithms for auto-encoders~\cite{hinton1994autoencoders}.
However, to the best of our knowledge, there is little published work on structural optimization in the deep learning community, with the notable exception of work based on empirical evaluation~\cite{bergstra2012random} and random search strategies~\cite{bergstra2012random}. Although, recently Bayesian optimization of hyper-parameters have been introduced \cite{snoek2012practical}. 

While these works are interesting, hyper-parameters are only one aspect of the DNN structure, albeit one which is closely related to the performance of the training algorithm. However, there are several other structural parameters that strongly affect DNN's performance which are usually set by experimental trial and error, such as network depth and for convolutional models the number of convolutional filters and kernel size for each layer. In our work, we aim to optimize model architecture, specifically targeting convolutional neural networks (CNNs), and optimizing complexity for each layer. Therefore, in our approach, the model architecture is not maintained constant during training, instead the model complexity is continuously optimized throughout the training step (parameter estimation by stochastic gradient descent), resulting, we believe, in a more scalable approach to the training of deep neural networks.
In section~\ref{framework}, we describe the general approach we are taking for problem of structure optimization of convolutional neural networks. In section~\ref{theory}, we describe the theoretical foundations of our approach. In section~\ref{results}, we discuss data-sets and experimental results and in section~\ref{discussion} we discuss about limitations and possible future improvements.

\section{Deep Clustered Convolutional Kernels}
\label{framework}
The basic idea for our Deep Clustered Convolutional Kernels (DCCKs) it a convolutional model architecture and associated structural training algorithm. 
We adopt a split/merge outer-loop to the training process that first increases model capacity to model new factors of variability seen in the data, then estimates new parameters for this larger model by stochastic gradient descent (SGD), and finally reduces model capacity to minimize model-space redundancy. Our approach takes inspiration by previous work in the area of Gaussian kernel HMMs~\cite{sankar1998experiments, rigazio2000optimal, bocchieri2001subspace, lee2001data}, and is philosophically based on Occam's razor principle whereby a smaller model with similar performance on a given data-set is likely to have better generalization capabilities to new unseen data.

An alternative view of work may be in the context of recent developments in DNN's compression: ~\cite{ba2014deep} shows that a (shallow) DNN can approach the performance of a substantially larger DNN when trained to mimic the logit output of the larger model. Similarly, ~\cite{gh2014dark} shows that logit-mimic training (referred to as ``Dark Knowledge'') results in orders of magnitude smaller models, compared to the initial complex ensemble models, yet provides competitive performance when tested on both small tasks (MNIST) as well as large scale industrial tasks (large vocabulary speech recognition).
It is important to notice that for both these works the authors acknowledge that, while such smaller high performance models can be obtained by logit mimic training from a more complex model set, thus showing that there is an optimal point in the parameter space with high performance, there is currently no known training procedure to directly achieve such optimal point in the smaller model. In this view finding such an elusive point in parameter space by systematically optimizing DNN's structure to eliminate redundancy and minimizing number of parameters, while at the same time estimating the model parameters under the given loss function. The main contribution of our work is a training methodology to iteratively optimize the number of convolutional kernels while estimating the convolutional filter parameters.

\subsection{Training algorithm}
\label{theory}

\begin{algorithm}[tb]
   \caption{Deep Clustered Convolutional Kernels training algorithm}
   \label{alg:1}
\begin{algorithmic}
   \STATE {\bfseries Input:} Initial network architecture $net$ with parameters $\lambda$, noise variance $\sigma_n$ and jitter angle $\sigma_{\alpha}$, stopping conditions $\delta_{0,1,2}$ and mini-batch size 
   \WHILE{$\Delta$ Validation Accuracy $> \delta_0$}
   \WHILE{$\Delta$ Validation Accuracy $> \delta_1$}
     \STATE // SPLIT
     \STATE $n_k$ = gaussianNoise($\sigma_n$)
     \STATE $\alpha_k$ = gaussianNoise($\sigma_{\alpha}$)
     \STATE $\lambda_1$ = $concat(\lambda, \lambda + n_k)$
     \STATE $\lambda$ = $concat(\lambda_1, rotate(kernel(\lambda), \alpha_k))$
     \STATE // FINETUNE
     \WHILE{$\Delta$ Validation Accuracy $> \delta_2$}
        \STATE runSGD($M$ minibatches)
     \ENDWHILE
   \ENDWHILE
     \STATE // MERGE
     \STATE $centroid$ = $Kmeans(kernels(\lambda))$
     \STATE $\lambda$ = $nearest(kernels(\lambda), centroid)$
     \WHILE{$\Delta$ Validation Accuracy $> \delta_2$}
        \STATE runSGD($M$ minibatches)
     \ENDWHILE
   \ENDWHILE
\end{algorithmic}
\end{algorithm}

Conceptually our training procedure is rather straightforward: starting from an initial network architecture, we first train the model by SGD until performance tops out on a validation set. Next, we increase the model complexity of selected convolutional layers by splitting the convolutional kernels. 
Splitting has the purpose of creating new plausible convolutional filters given the current set of filters and can be done by applying image pre-processing techniques to the kernels, as well as adding jittering and noise to create enough variation. 
After splitting, the model is again trained by SGD and possibly split again until performance tops out. At that point model is merged to reduce redundancy in the parameter space and again trained by SGD. 
Notice that the split/merge procedure can start at any layer but than has to propagate upwards to change the number of kernels of the connecting layers (fan-out). In our setup, given by input data $x$, forward propagation $f$ is done by:

\begin{equation}
\label{eq1}
f(x) = g(Wx + B)
\end{equation}
where $g$ is ReLU activation function with $g(x) = max(0, x)$, $W$ is the weight parameters of the convolutional layer, and $B$ is the biases, each with the following dimensions:
\begin{equation}
\label{eq2}
 W_l = N_l \times \overbrace{d_l \times k_l \times k_l}^{P}
\end{equation}

\begin{equation}
\label{eq3}
 B_l = 1 \times 1 \times 1 \times N_l
\end{equation}

where $l$ is a convolutional layer with $l \in \{1,...,L\}$, $N_l$ is the number of outputs of $l$, $d_l$ is number of channels of $l$, and $k_l$ is size of kernel used for $l$. We use square convolutional kernels, so kernel dimensions are $k_l \times k_l$. For simplicity, we define  sub-dimension of W as $P$, shown in \ref{eq2}. For the first convolutional layer we have $d_1 = 3$ for RGB images and $d_1 = 1$ for gray-scale images. In the following convolutional layers, $d$ is the output of the previous convolutional layer thus, $P$ would be the size of the feature vectors. This implies that, when we perform the split/merge steps for level $l$, we need to update both $W_l$ and $B_l$ as well as $W_{l+1}$. Biases for the following convolutional layer are independent.
An important caveat is that the order of the optimal split/merge operation depends on the specific data-set and the filter parameters. For instance, if the initial filters are sparse it is beneficial to do merge first. Otherwise, it is best to perform split first especially on smaller data-sets when the initial filters are already compact and discriminative.

\subsubsection{Splitting Kernels}
\label{split-kernel}

\begin{figure}[ht]
\vskip 0.2in
\begin{center}
\centerline{\includegraphics[clip,scale=0.7, trim=1.2cm 1cm 0.2cm 0.5cm]{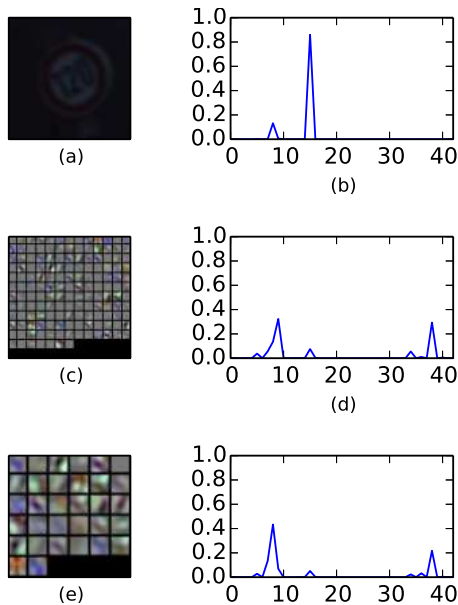}}
\caption{(a) Highly tilted, misclassified test image (b) Soft-max output of original baseline model resulting in miss-classification (c) Baseline model convolutional kernels: notice high proportion of redundant kernels (d) Soft-max at DCCKs intermediate training stage, after split and fine-tuning (e) Final DCCK convolutional kernels, after merge and fine-tuning, showing reduced redundancy (f) Final DCCK soft-max output, correctly classifying the image}
\label{fig:splittingTSR}
\end{center}
\vskip -0.2in
\end{figure}

With splitting, we want to increase model complexity by creating new convolutional kernels from the set of existing well-trained kernels. Therefore, we create new kernels by selectively choosing from a fixed set of transformations. The possible set of transformations to play with is vast and includes the six isometries of the plane, angular rotation, change in contrast (negative ``reversing'') and many others. In our experiments, we focus on two transformations that seemed to provide a consistent improvement:
\begin{itemize}
	\item \textbf{Rotation} creates new kernels by rotating existing kernels in random directions.
	\item \textbf{Noise perturbation} creates new kernels by adding Gaussian noise to the existing kernels.
\end{itemize}
One important aspect we verified in our experiments is that rotating kernels has a lower computational cost at training time than rotating training images to create augmented training set. Moreover, we observed that rotating the filters can help improve robustness for highly tilted objects outlets, which would be otherwise hard to correctly classify (see Figure~\ref{fig:splittingTSR}).
Adding random Gaussian noise, on the other hand, has the obvious benefit of creating diversity and helping with the SGD, like previously reported by~\cite{JMLR:v15:srivastava14a}.
Regarding the splitting strategy, currently we took the simplest approach and split every kernel by a fixed amount. This is bound to be locally unoptimal, and surely a better splitting strategy that tries to maximize some diversity or discrimination criteria could be devised, instead of indiscriminately splitting every single kernel. However, for the most part, we observe that wasteful parameters created by this simple splitting strategy will be eliminated during the final merging step; therefore, aside from a potential sub-optimality in the CPU/Memory usage, we speculate the final model accuracy might not be very affected by this uniform splitting strategy.

\subsubsection{Merging Kernels}
\begin{figure}[ht]
\label{fig:splittingTSR1}
\vskip 0.2in
\begin{center}
\begin{tikzpicture}[line/.style={->,shorten >=0.4cm,shorten <=0.4cm},ultra thick]
\usetikzlibrary{decorations.pathmorphing}

\coordinate (I1) at (1.2,6.4);
\coordinate (I2) at (5.5,4);
\coordinate (I3) at (1.4,2.0);
\coordinate (T1) at (1.3,8.2);
\coordinate (T2) at (1.5,-0.4);
\coordinate (T3) at (5.6,0.4);
\coordinate (T4) at (5.6,2.2);
\node (P1) [inner sep=0pt,minimum width=2cm]  at (I1) 
           {\includegraphics[scale=0.23]{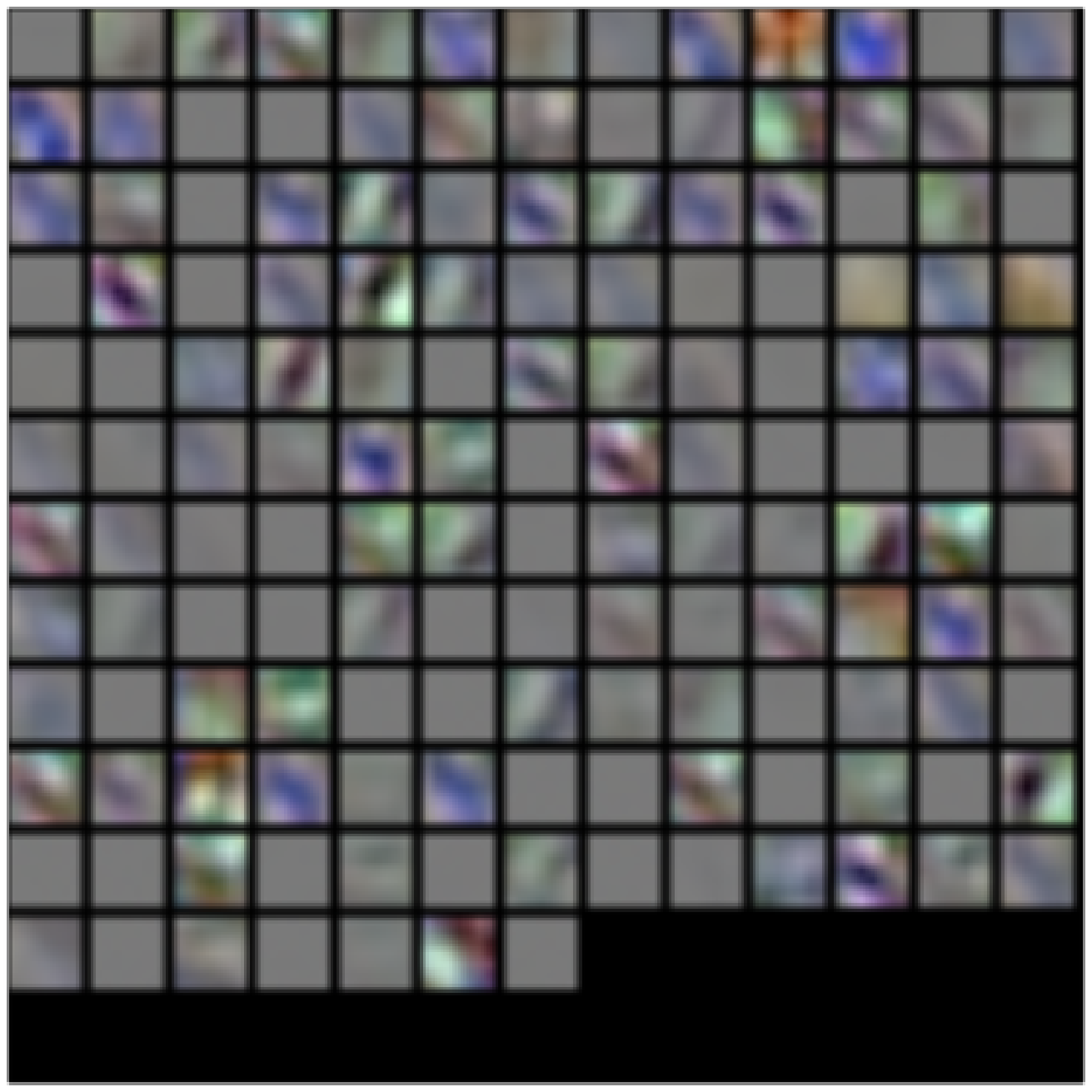}};
\node (P2) [inner sep=0pt,minimum width=1.5cm] at (I2)
           {\includegraphics[scale=0.22]{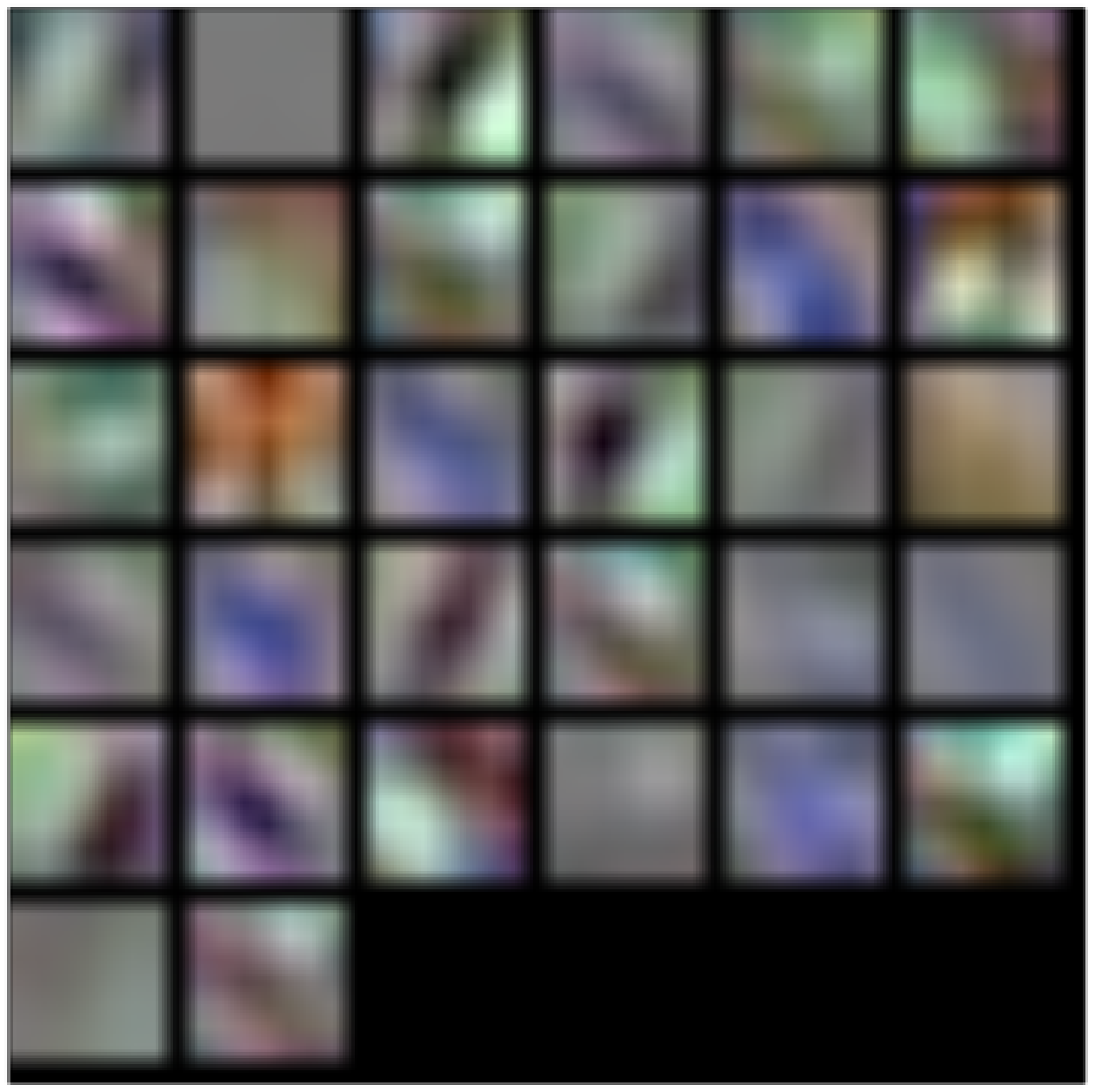}};
\node (P3) [inner sep=0pt,minimum width=1cm] at (I3) 
           {\includegraphics[scale=0.27]{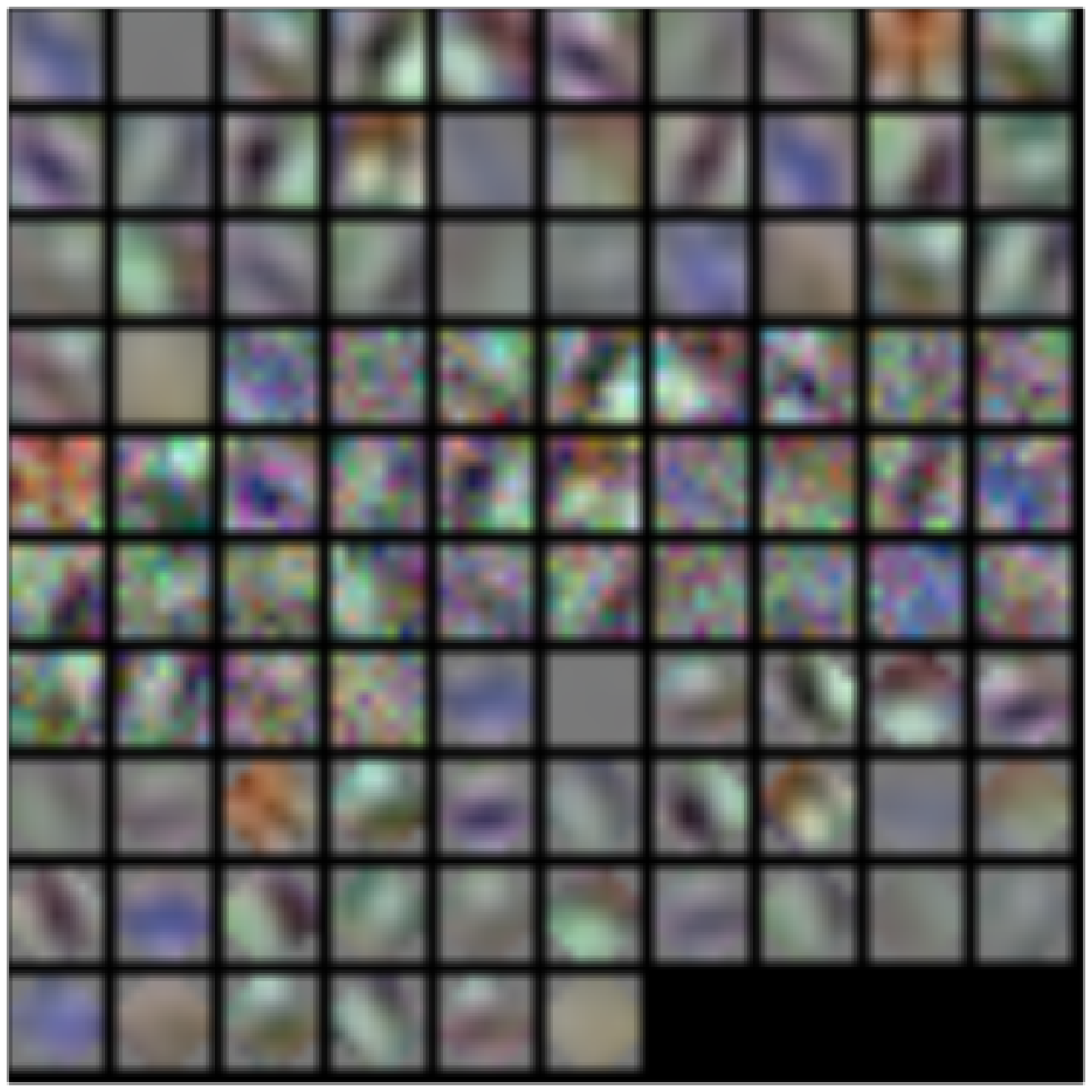}};
\node (F1) [text width=1.3cm, decorate, decoration=saw, fill=yellow!10,draw,circle,thick] at (T3)
      {\textbf{Finetune}};
\node (F2) [text width=1.6cm,rectangle, thick] at (T2)
      {\textbf{96 kernels}};
\node (F3) [text width=1.9cm,rectangle, thick] at (T1)
      {\textbf{150 kernels}};
\node (F4) [text width=1.6cm,rectangle, thick] at (T4)
      {\textbf{32 kernels}};
\draw[->] ([xshift=-7pt]P1.east) -- ([yshift=-10pt]P2.north)
    node[midway,fill=white] {\textit{MERGE}};
\draw[->] ([yshift=-2pt]P2.south) -- ([yshift=4pt]F1.north);
\draw[<-] ([xshift=-10pt]P3.east) -- ([xshift=-4pt]F1.west)
    node[midway, fill=white] {\textit{SPLIT}};
\path [black,line,out=280,in=200] (P3) edge (F1);
\end{tikzpicture}
\caption{DCCK training example: starting from a large GTSRB model 150 convolutional kernels for the first layer, the algorithm first merges it to 32 kernels. After fine-tuning, kernels are split by adding noise and rotating, then fine-tuned one more time.}
\end{center}
\vskip -0.2in
\end{figure}

After the splitting step, the model might have too much capacity and thus part of the model might become over-parameterized, possibly resulting in over-fitting and lower generalization power.
Therefore the merging step has the purpose of removing model space redundancies and reducing model size, while maintaining the overall model accuracy. In our algorithm we use \textit{k}-means clustering to merge kernels since, naturally, \textit{k}-means cluster distortion under the defined distortion measure (we employ $L2$ norm to compute cluster distortion).
We empirically observe that \textit{k}-means clustering to merge filter maps is effective in reducing kernel's redundancy (see filters in~\ref{fig:splittingTSR1}).
Then, we train the network and get weight and bias matrices from each convolutional layer to then choose the filters that are nearest to each centroid. We update $W_l$ and $B_l$, with $W_l'$ and $B_l'$ using $k$-means clustering to get centroids $C$ as:

\begin{equation}
\label{eq4}
C = \underset{P}{\arg\min} \overset{\mathcal{C}}{\underset{j=1}{\sum}} \ \underset{p \in P}{\sum} || p - \mu_j ||^2
\end{equation}

\begin{equation}
\label{eq5}
W'_l = \left\{
\begin{array}{ll}
[P'_1, ..., P'_i, ...,  P'_\mathcal{C}] , \ or \\

[C_1, ..., C_i, ...,  C_\mathcal{C}]
\end{array}
\right.
\end{equation}

where

\begin{equation}
\label{eq6}
P'_i = \underset{P'}{\arg\min} \ ||P' - C_i \ ||^2 \ \ , \ \  i = \{1,...,\mathcal{C}\}
\end{equation}

and finally,

\begin{equation}
\label{eq7}
B'_l = \left\{
\begin{array}{ll}
[B'_1, ..., B'_i, ...,  B'_\mathcal{C}] , \ or \\

[{\beta}_1, ..., {\beta}_i, ..., {\beta}_\mathcal{C}]
\end{array}
\right.
\end{equation}

where $B'_i$ is $P'_i$'s matched biases matrix, and ${\beta}_i$ is 

\begin{equation}
\label{eq8}
{\beta}'_i = \frac{\underset{i}{\sum}\ b_n}{\mathcal{\eta}_i} \ \ , \ \   n = \{1,...N_l\}
\end{equation}
where
$\eta_i$ is number of $p$ in group $C_i$.

As shown in \ref{eq5} and \ref{eq8}, we explored two different methods to update $W$ and $B$. The first method consists in choosing the $Pi$ that is closer to each centroid $C_i$. In this case, we use the correspondent bias vector $B_i$ to the corresponding $P_i$ selected. The other way is to use the centroid $C_i$ itself as filter parameters and update $B_i$ with average bias from each cluster.
An important detail to choose the right value of \textit{k}: if we choose \textit{k} too small then average cluster distortion will be too high to appropriately represent the model parameters, possibly resulting in ineffective features maps. On the other hand, if we choose \textit{k} too big, not enough kernels will be merged. This, unfortunately, may very well be a hyper-parameter that will have to be manually tuned. Table~\ref{table-MNIST_Network} shows \textit{k} selected for each experiment which gives the best results on our network models.

\section{Experimental results}
\label{results}
Our experimental results are based on three different data-sets: MNIST, German Traffic Sign Recognition Benchmark (GTSRB), and CIFAR-10.
As much as possible, to make our experiments significant and to validate our approach, we started from hand-tuned model architectures that were as close as possible to the state of the art, in an effort to prove that our split/merge training procedure can still improve model architecture even when starting from a very highly tuned architecture. Baseline performance are reported in Table~\ref{table-orig}. For all experiments, we used the BVLC Caffe C++ package~\cite{jia2014caffe}.
We started our experiments from MNIST since the quick training time allowed to quickly  determine reasonable range of hyper-parameters such as the number of centroids \textit{k}, number of kernels for the split/merge procedure. Next, we move to a more realistic task such as GTSRB for which we started from an initial model, extremely close to the state of the art and finally confirm the portability of our findings on the harder CIFAR-10 data-set. We report the details of each data-set experiments in the following sections.

\begin{table}[t]
\vskip 0.15in
\begin{center}
\begin{small}
\begin{sc}
\begin{tabular}{lcccr}
\hline
\abovespace\belowspace
MNIST & GTSRB1 & GTSRB-3DNN & CIFAR-10 \\
\hline
\abovespace
0.82\% & 2.44\% & 1.24\% & 10.4\% \\
\hline
\end{tabular}
\end{sc}
\end{small}
\end{center}
\vskip -0.1in
\caption{Baseline models performance on the three data-sets selected for our experiments.}
\label{table-orig}
\end{table}

\subsection{MNIST results}
\label{mnist-results}
The MNIST data-set contains 60,000 training images and 10,000 testing images of hand-written digits of size 28x28.
The baseline model is composed of two convolutional layers and two fully-connected layers, as shown in Table~\ref{table-MNIST_Network}, with ReLU and pooling following each convolutional layer. This baseline model achieves 0.82\% Error rate with this simple network. The DCCKs training algorithm begins by splitting the first convolutional layer from 100 to 200 kernels; after the subsequent fine-tuning the model achieved 0.59\% error rate, which is almost 30\% relative improvement from the original model. 
This compared favorably to a 200 kernel models trained from scratch, which achieves 0.78\%, and even a 300 kernels model trained from scratch, which achieves  0.75\%. This verifies that splitting filters has the potential to help the following SGD based fine-tuning to achieve an optimal point which generalists better. Also, more importantly after following merging step, back to 100 kernels, the performance dropped only 0.01\% to an error rate of 0.59\%.

\begin{table}[t]
\vskip 0.15in
\begin{center}
\begin{small}
\begin{sc}
\begin{tabular}{lcccr}
Layer & \# of maps & Kernel \\
\hline
\abovespace
Input           & 3   &      \\
Convolutional   & 100 & 5x5  \\
Max Pooling     & 100 & 2x2  \\
Convolutional   & 50  & 5x5  \\
Max Pooling     & 50  & 2x2  \\
Fully connected & 100 & 1x1  \\
Fully connected & 10  & 1x1  \\
\hline
\end{tabular}
\end{sc}
\end{small}
\end{center}
\vskip -0.1in
\caption{MNIST baseline architecture}
\label{table-MNIST_Network}
\end{table}

\begin{table}[t]
\vskip 0.15in
\begin{center}
\begin{small}
\begin{sc}
\begin{tabular}{lcccr}
\hline
\abovespace\belowspace
No. & stage & conv1 & conv2 & Err(\%) \\
\hline
\abovespace
1 & original       & 100 & 50 & 0.82    \\
2 & original       & 200 & 50 & 0.78    \\
3 & original       & 300 & 50 & 0.75    \\
4 & split from [1] & \textbf{200} & 50 & 0.58    \\
5 & merge from [4] & \textbf{100} & 50 & \textbf{0.59}    \\
\hline
\end{tabular}
\end{sc}
\end{small}
\end{center}
\vskip -0.1in
\caption{MNIST error rate after fine-tuning. Notice that clustering was performed on the first convolutional layer only.}
\label{table-MNIST}
\end{table}

\subsection{GTSRB results}
\label{gtsrb-results}

\begin{table}[t]
\vskip 0.15in
\begin{center}
\begin{small}
\begin{sc}
\begin{tabular}{lccr}
Layer & \# of Maps & Kernel(GTSRB-3DNN) \\
\hline
\abovespace
Input           & 3   & \\
Convolutional   & 150 & 3x3, 3x3, 3x3 \\
Max Pooling     & 150 & 2x2, 2x2, 2x2 \\
Convolutional   & 150 & 4x4, 4x4, 2x2 \\
Max Pooling     & 150 & 2x2, 2x2, 2x2 \\
Convolutional   & 250 & 4x4  4x4, 2x2 \\
Max Pooling     & 250 & 2x2  2x2, 2x2 \\
Fully Connected & 500 & 1x1  1x1, 1x1 \\
Fully Connected & 43  & 1x1  1x1, 1x1\\
\hline
\end{tabular}
\end{sc}
\end{small}
\end{center}
\vskip -0.1in
\caption{GTSRB-3DNN architecture}
\label{table-MCDNN-Network}
\end{table}

\begin{table}[t]
\vskip 0.15in
\begin{center}
\begin{small}
\begin{sc}
\begin{tabular}{lccccr}
\hline
\abovespace\belowspace
No. & stage & conv1 & conv2 & conv3 & Err(\%) \\
\hline
\abovespace
1  & original  & 150          & 150           & 250          & 2.44 \\
2  & merge [1] &  \textbf{32} & 150           & 250          & 2.34 \\
3  & merge [2] &  32          & \textbf{32}   & 250          & 2.7  \\
4  & merge [2] &  32          & \textbf{64}   & 250          & 2.36 \\
5  & merge [3] &  32          &  32           &  \textbf{32} & 3.82 \\
6  & split [2] &  \textbf{64} & 150           & 250          & 2.5 \\
7n & split [3] &  32          & \textbf{64}   & 250          & 2.25 \\
8r & split [3] &  32          & \textbf{64}   & 250          & \textbf{2.15} \\
9  & split [1] & \textbf{300} & 150           & 250          & 2.24 \\
10 & merge [1] & \textbf{40}  & 150           & 250          & 2.31 \\
11 & split [1] &  150         & \textbf{300}  & 250          & 2.27 \\
\hline
\end{tabular}
\end{sc}
\end{small}
\end{center}
\vskip -0.1in
\caption{GTSRB1 baseline model experiments, 'R' denotes 'Rotation', and 'N' denotes 'Noise perturbation'. Remark that [8R] which splits both the first and the second convolutional layer, followed by merge of the second layer, achieved the best performance. Instead, [5] which merges the last convolutional layer had a performance drop; we speculate that this is due to difficulty in optimizing the following fully connected layer.}
\label{table-GTSRB1}
\end{table}

\begin{table}[t]
\vskip 0.15in
\begin{center}
\begin{small}
\begin{sc}
\begin{tabular}{lcccrl}
\hline
\abovespace\belowspace
No. & stage & conv1 & conv2 & conv3 & Err(\%) \\
\hline
\abovespace
1  & original  & 150          & 150           & 250          & 1.24 \\
2  & original  & 16           & 150           & 250          & 1.67 \\
3  & merge [1] & \textbf{32}  & 150           & 250          & \textbf{1.18} \\
4  & merge [1] & \textbf{16}  & 150           & 250          & 1.25 \\
5  & split [1] & \textbf{300} & 150           & 250          & 1.21 \\
6  & split [3] & \textbf{64}  & 150           & 250          & \textbf{1.15} \\
\hline
\end{tabular}
\end{sc}
\end{small}
\end{center}
\vskip -0.1in
\caption{Results table for DCCK trained from the state of the art GTSRB-3DNN initial model, showing a small but significant improvement.}
\label{table-GTSRB-3DNNs}
\end{table}

The GTSRB data-set contains 39,209 training images and 12630 testing images of various size, with 43 different classes consisting of standard traffic signs from Germany~\cite{Houben-IJCNN-2013}. First, we resized all images to 48x48 and then we applied pre-processing techniques such as histogram equalization, adaptive histogram equalization, and contrast normalization.
For this task, we have two sets of initial networks: a single model baseline GTSRB1, consisting of three convolutional and two fully connected reaching 2.44\% error rate, and larger state of the art ensemble model GTSRB-3DNN (Table~\ref{table-MCDNN-Network}), inspired by MCDNN\cite{Ciresan12multi-columndeep}, and reaching 1.24\% error rate, which is within 0.2\$ from the best published result. We remark the ensemble models use different input size of 48x48 pixels, 38x48 pixels and 28x48 pixels: because of this, we expected a high degree of redundancy on the GTSRB-3DNN kernels which may be successfully exploited by the DCCKs merging step. 
Indeed, by visually inspecting the lower convolutional layers we could easily identify an abundant amount of redundancy (see~\ref{fig:splittingTSR1}). 
Because of this highly redundant structure in the initial model, we inverted the sequence of our training procedure to first merge kernels instead of splitting, which maintains the accuracy and provides significantly faster training~\ref{table-GTSRB-speed}.

Furthermore, the specific structure of the traffic signs provided for some peculiar behaviors on this database: for instance, kernel rotation especially helped improving performance. A detailed inspection of the recognition errors highlighted that several traffic signs were misclassified by the baseline model were highly tilted; such instances were mostly recovered and correctly recognized after DCCKs training (see Figure~\ref{fig:splittingTSR} for one example of such instance). We also remark that using centroids as new kernels resulted in better gains on this data set.

Table~\ref{table-GTSRB1} and Table~\ref{table-GTSRB-3DNNs} shows the experimental results. We remark that in all the experiments, in almost all cases, we either achieve significantly better performance or similar performance with significantly reduced model size. One exception worth noticing is [5] in Table~\ref{table-GTSRB1} which shows the worst performance of all experiments: in this case we  merged the last convolutional layer which is fully connected to the first fully connected layer of this network architecture. We speculate this issue is due to the fact that is notoriously hard to optimize parameters of fully connected layers, splitting a convolutional layer which fans-out into a fully connected layer has the potential to harm the parameter structure to a point where SGD cannot easily recover.

\subsection{CIFAR10 results}
\label{cifar10-results}

\begin{table}[t]
\vskip 0.15in
\begin{center}
\begin{small}
\begin{sc}
\begin{tabular}{lcccrl}
\hline
\abovespace\belowspace
No. & stage & conv1 & conv2 & conv3 & Err(\%) \\
\hline
\abovespace
1  & original  & 192          & 192           & 192          & 10.4  \\
2  & split [1] & \textbf{384} & 192           & 192          & 10.29 \\
3  & split [1] & \textbf{576} & 192           & 192          & 10.25 \\
4  & merge [3] & \textbf{192} & 192           & 192          & \textbf{10.2}  \\ 
5  & split [1] & 192          & 192           & \textbf{384} & 10.04 \\
6  & split [1] & 192          & \textbf{384}  & 192          & 10.04 \\
7  & merge [6] & 192          & \textbf{192}  & 192          & 10.28 \\
\hline
\end{tabular}
\end{sc}
\end{small}
\end{center}
\vskip -0.1in
\caption{Result table for CIFAR-10.}
\label{table-CIFAR10}
\end{table}

The CIFAR-10 data-set consists of 50,000 training and 10,000 testing images. Each image is 32x32 pixels and represent a class of natural occurring objects. To develop the CIFAR-10 baseline we used the same techniques discussed in~\citep{Goodfellow_maxout_2013} and the Network-In-Network~\citep{NIN} model which achieves a baseline 10.4\% error rate, which is within reasonable distance from to the state of the art. When we apply DCCKs training on the CIFAR-10 data-set, the increased performance is not as large as on the previous data-sets but it is still significant and consistent. We believe that this is because the highly successful highly (manually) optimized Network-In-Network architecture makes it harder for the automatically devised DCCKs training to provide a large improvement. Therefore these results should demonstrate that DCCKs may still provide some improvement even when applied on top of more complex highly tuned architectures, while keeping the number of parameters under control.
Additionally we show that by splitting layers and doubling the number of parameters we could achieve an additional 0.5\%  average error rate improvement.

\begin{figure}[ht]
\vskip 0.2in
\begin{center}
\centerline{\includegraphics[clip,scale=0.3, trim=0cm 1cm 0cm 2.5cm]{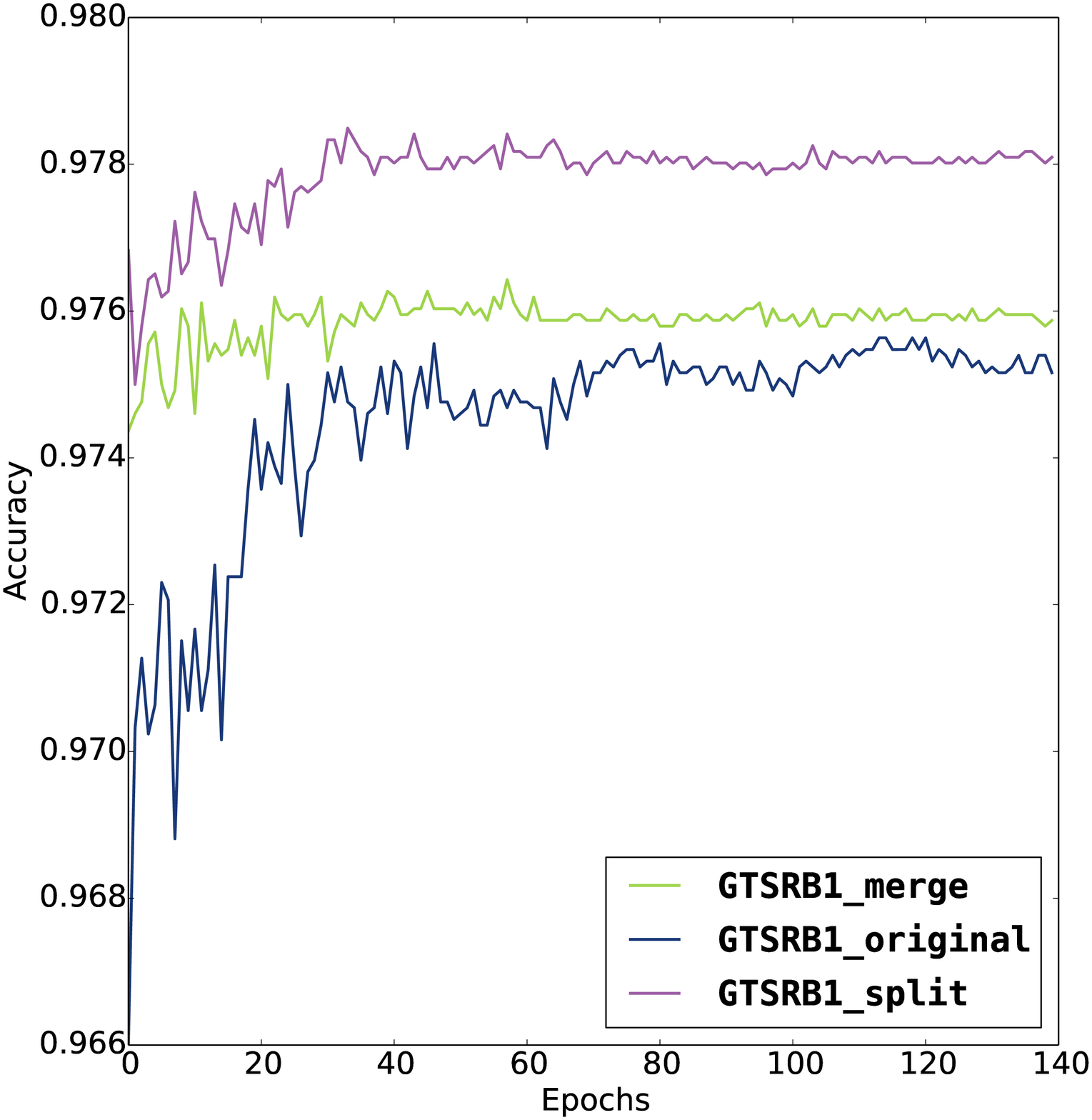}}
\caption{Test-set accuracy of GTSRB1 (simple) network during fine-tuning. Notice that GTSRB1\_merge and GTSRB1\_merge have the same number of parameters, but the optimized DCCK architecture shows better accuracy throughout epochs.}
\label{fig:plotTSR}
\end{center}
\vskip -0.2in
\end{figure} 

\begin{figure}[ht]
\vskip 0.2in
\begin{center}
\centerline{\includegraphics[clip,scale=0.3, trim=0cm 1cm 0cm 2.5cm]{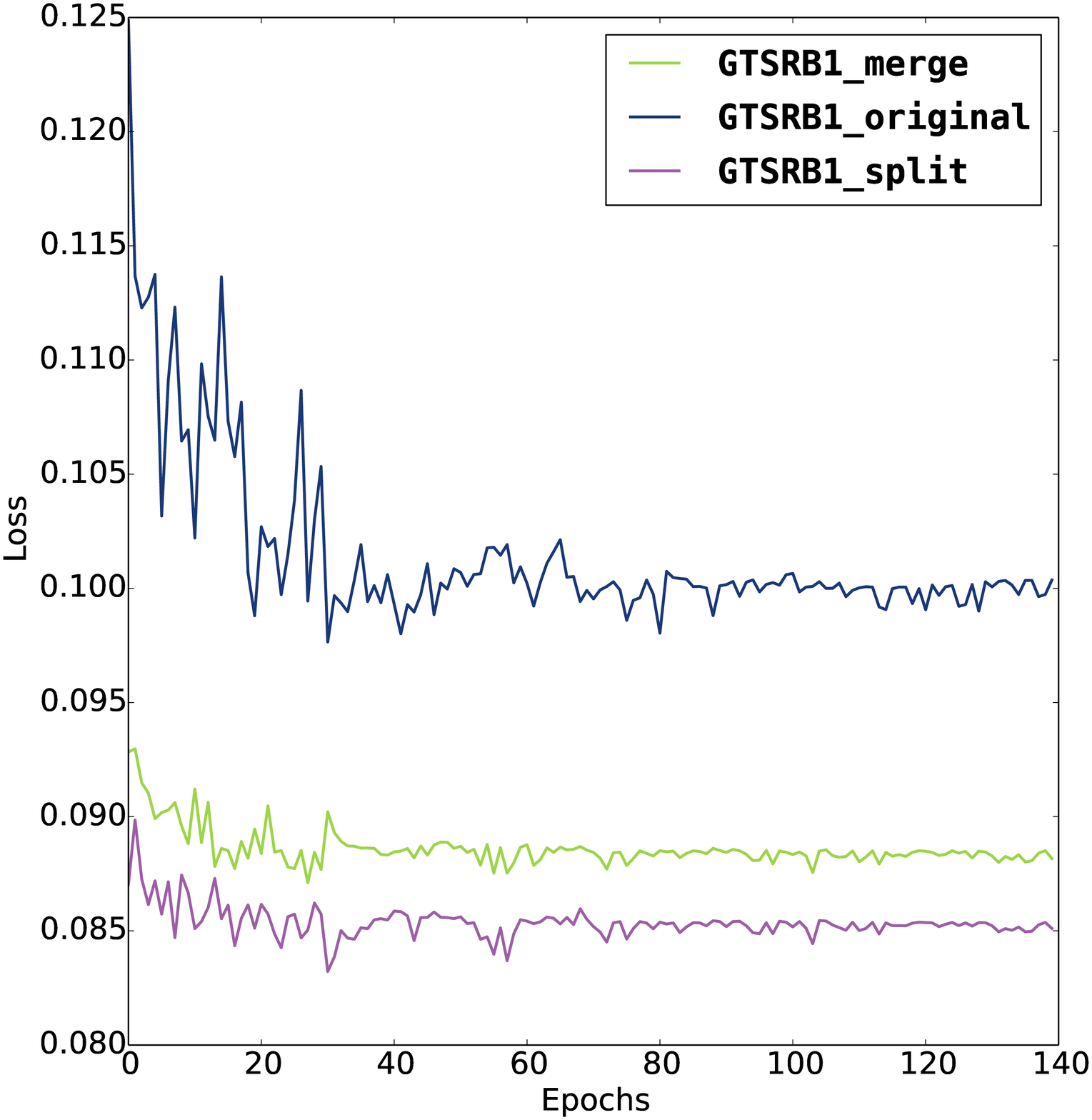}}
\caption{Test-set loss of GTSRB1 (simple) network during fine-tuning. Notice that GTSRB1\_merge and GTSRB1\_merge have the same number of parameters, but the optimized DCCK architecture shows better accuracy throughout epochs.}
\label{fig:plotTSR1}
\end{center}
\vskip -0.2in
\end{figure} 

\begin{table}[t]
\vskip 0.15in
\begin{center}
\begin{small}
\begin{sc}
\begin{tabular}{lcccrl}
\hline
\abovespace\belowspace
Model & stage & conv1 & conv2 & speed(ms) \\
\hline
\abovespace
1. simple  & original  & 150  & 150 & 14.8  \\
2. simple  & merge [1] & 32  &  150 & 14.1  \\
3. simple & merge [2] & 32  &  64 &   12.6  \\
4. 3-DNNs   & original  & 150 & 150 & 27.9 \\
5. 3-DNNs   & merge [4]  & 32 & 150 & 19.4 \\
\hline
\end{tabular}
\end{sc}
\end{small}
\end{center}
\vskip -0.1in
\caption{Speed comparisons for GTSRB1 and GTSRB-3DNNs models and their corresponding DCCK trained models. Test time of forward-pass only with minibatches of 10 48x48 pixel images on nVidia GeForce GTX 770.}
\label{table-GTSRB-speed}
\end{table}

\section{Discussion}
\label{discussion}
In this work, we introduced the concept of DCCKs and introduced a training procedure whereby convolutional kernels learned by SGD can be effectively split and merged. Experimental results confirmed this process results in gradually improving performance, while the training algorithm jointly optimizes structure as well as model's parameters. Results show that DCCKs can make parsimonious use of model capacity by converging towards the minimal number of parameters that gives the best performance, even when starting with highly manually optimizing network architecture. Figure~\ref{fig:plotTSR} shows training and validation data loss over fine-tune epochs; the ``original'' and the ``merge'' curves refer to training and generalization loss for models having the same number of parameters; notice how the ``merge'' curve is consistently above the ``original'' curve, apparently providing an upper-bound to the loss, and thus empirically confirming that the DCCKs architecture was indeed an improved by the training algorithm. Moreover, in some experiments, DCCKs resulted in significantly higher performance with smaller number of parameters than the original model. 
On the other hand, DCCKs showed bigger gains on simpler databases, such as MNIST and GTSRB, than on more complex CIFAR-10 data-set and to the more complex Network-In-Network model architecture. This is however to be expected, especially because the NIN architecture is extremely well tuned and very high performance to begin with, so it is natural to expect smaller gains by our automatic structure optimization procedure.
Beside the obvious advantage of automatic structure optimization, a side benefit of DCCKs training is that manipulating kernels takes less computations than pre-processing training data, which makes DCCKs optimization more efficient.

To conclude, we believe there are several aspects of DCCKs training algorithm that could be improved. As we mentioned in~\ref{split-kernel} currently all kernels are split by the same amount. However, one could argue that some kernels might be better than others and should be replicated first, possibly based on the ability provide new discriminative features. If we could determine such kernels we could potentially improve training speed, though, potentially final accuracy after the merge step might not be much impacted as much. Finding a more extensive set of kernel transformations to achieve a highly selective split step would also be an appropriate next step, as well as comparison and combination with logit-mimic training and model compression techniques  ~\cite{ba2014deep, gh2014dark}. Ultimately, like for any new methodology in the deep-learning sector, it would be very important to test how well DCCKs scale  higher dimensional larger problems, such as IMAGE-NET and to different non-vision tasks such as speech recognition or language modeling.

\bibliography{ref}

\bibliographystyle{icml2015}

\end{document}